# Climate Knowledge in Large Language Models

By Ivan Kuznetsov[1]*, Jacopo Grassi[2], Dmitrii Pantiukhin[1], Boris Shapkin[1], Thomas Jung[1,3], Nikolay Koldunov[1]

[1] Alfred Wegener Institute, Helmholtz Centre for Polar and Marine Research, Bremerhaven, Germany.
[2] Department of Environment, Land, and Infrastructure Engineering, Politecnico di Torino, Turin, Italy.
[3] Institute of Environmental Physics, University of Bremen, Bremen, Germany.

## Abstract

Large language models (LLMs) are increasingly deployed for climate-related applications, where understanding internal climatological knowledge is crucial for reliability and misinformation risk assessment. Despite growing adoption, the capacity of LLMs to recall climate normals from parametric knowledge remains largely uncharacterized. We investigate the capacity of contemporary LLMs to recall climate normals without external retrieval, focusing on a prototypical query: mean July 2-m air temperature 1991-2020 at specified locations. We construct a global grid of queries at 1° resolution land points, providing coordinates and location descriptors, and validate responses against ERA5 reanalysis. Results show that LLMs encode non-trivial climate structure, capturing latitudinal and topographic patterns, with root-mean-square errors of 3-6 °C and biases of ±1 °C. However, spatially coherent errors remain, particularly in mountains and high latitudes. Performance degrades sharply above 1500 m, where RMSE reaches 5-13 °C compared to 2-4 °C at lower elevations. We find that including geographic context (country, city, region) reduces errors by 27% on average, with larger models being most sensitive to location descriptors. While models capture the global mean magnitude of observed warming between 1950-1974 and 2000-2024, they fail to reproduce spatial patterns of temperature change, which directly relate to assessing climate change. This limitation highlights that while LLMs may capture present-day climate distributions, they struggle to represent the regional and local expression of long-term shifts in temperature essential for understanding climate dynamics. Our evaluation framework provides a reproducible benchmark for quantifying parametric climate knowledge in LLMs and complements existing climate communication assessments.

## Introduction

Large language models (LLMs) have demonstrated substantial capacity to store and recall factual information acquired during pretraining [1,2,3]. As these systems are increasingly deployed as assistants in scientific domains, understanding the scope and reliability of their internal knowledge becomes critical. In climate science, users frequently seek factual information about temperature, precipitation, or other variables at specific locations and time periods. Modern climate applications of LLMs have begun to address this through retrieval-augmented generation, where external authoritative sources ground model responses [4,5,6,7]. For instance, ChatClimate integrates content from IPCC reports to answer climate questions [4], while ClimSight combines LLM reasoning with climate model databases to provide localized projections [5,6]. These tool-augmented systems underscore the value of external knowledge - yet many real-world interactions with LLMs occur in closed-book (no retrieval of external data) settings where the model relies solely on parametric knowledge.

This raises a fundamental question: what climatological information does an LLM implicitly encode from its training data, and how reliable is it? Recent work has examined LLMs' factual recall capabilities across diverse domains [1,2,3,8,9,10,11]. Studies show that while LLMs can retrieve many facts stored in their parameters, accuracy degrades for rare or fine-grained information, the so-called "long tail" of knowledge [3,11]. Models also exhibit challenges in maintaining factual consistency [8,9] and can hallucinate when uncertain [10]. However, existing evaluations of LLMs' geoscientific knowledge focus on general geographical facts (e.g., populations, elevations) or climate information retrieved via external sources; to our knowledge, no prior work has systematically assessed LLMs' ability to recall location-specific numeric climate data, such as climate normals, from their parametric memory.

We investigate this question by asking contemporary LLMs a simple, prototypical query: "What is the mean July 2-m air temperature at location (latitude, longitude) for the period 1991-2020?" This query mimics a common request to a climate data archive but is formulated in natural language to elicit a single numeric answer. By posing such questions at 15,395 land locations on a global 1° grid (ocean points excluded) and comparing responses to reanalysis data, we can assess the extent to which these models have absorbed climate norms during pretraining. This constitutes a closed-book climatology challenge where each location has an unambiguous correct value that can be verified against reference datasets. We use ERA5 reanalysis (1991-2020) as our ground truth [22].

The motivation is both scientific and practical. If LLMs encode substantial climate statistics, they might correctly answer common climate queries without external tools. If they do not, any confident answers could be misleading. Understanding this balance is important for deploying LLMs in climate-related applications where misinformation carries high stakes. By isolating the LLM in closed-book mode (no tools or retrieval allowed), our evaluation specifically probes parametric knowledge, the information about climate stored in internal model weights.

Numerous frameworks now evaluate LLMs on diverse capabilities [12,13,14,15], with recent efforts focusing on scientific domains [16,17]. However, climate-specific evaluations have primarily assessed textual understanding and communication [7,18] rather than quantitative recall of climate variables. Our work complements these by requiring exact numeric answers for objectively verifiable facts, enabling straightforward spatial error analysis. While our demonstration focuses on July temperature, the approach extends naturally to other months, variables (e.g., precipitation), or future scenarios. We provide this as a reproducible evaluation framework that others can adopt and extend.

Our evaluation offers a new lens to examine LLMs on quantitative scientific knowledge. We emphasize that our goal is to characterize what the model knows, not to claim any predictive capability or physical understanding. The framework facilitates objective comparisons across models and can track progress as newer models are released or as domain-specific models are fine-tuned on climate texts [18].

## Results

We evaluated 17 model configurations across multiple model families on the July temperature task (Table 1). For each location, we made 10 independent queries to each model and used the mean of these responses for final evaluation. This approach accounts for sampling variability and provides robust estimates of model predictions.

We constructed a global 1° land grid, generated standardized natural language prompts (Box 1), and queried models in zero-shot mode, validating outputs against ERA5 reanalysis [22]. Our evaluation reveals that current LLMs contain a non-trivial representation of Earth's climate, though with significant quantitative limitations. Model predictions captured the broad expected structure of Earth's climate: warmer tropics and hot subtropical deserts, cooler high latitudes and mountain regions (Figure 1). Visual comparison between ERA5 observations (Figure 1a) and a representative model output (Figure 1b, GPT-5 with sampling temperature 0.3) reveals qualitative agreement in large-scale patterns. However, quantitative metrics reveal substantial deviations. Global root-mean-square errors (RMSE) ranged from 3.49 °C (GPT-5, temp=0.3) to 10.20 °C (eci-io-climategpt 70b), with a median near 5 °C across models (Table 1); most models fell within 3-6 °C. Mean biases were on the order of ±1 °C, with spatially coherent error structures. Interestingly, these error magnitudes are comparable to biases in contemporary physical climate models when evaluated against reanalysis at similar spatial scales [19], though we stress that this comparison is purely contextual - LLMs lack physical understanding, and their errors arise from fundamentally different mechanisms.The best-performing models showed mean biases ranging from -0.57 °C to +0.88 °C, while the full ensemble spanned from -1.37 °C to +3.93 °C. Importantly, individual models maintained consistent bias sign across different configurations, whether changing sampling parameters or adding or removing location information (country, city, region). This indicates that the systematic tendency to under-predict or over-predict temperatures is an intrinsic characteristic of each model rather than an artifact of specific prompt variations.

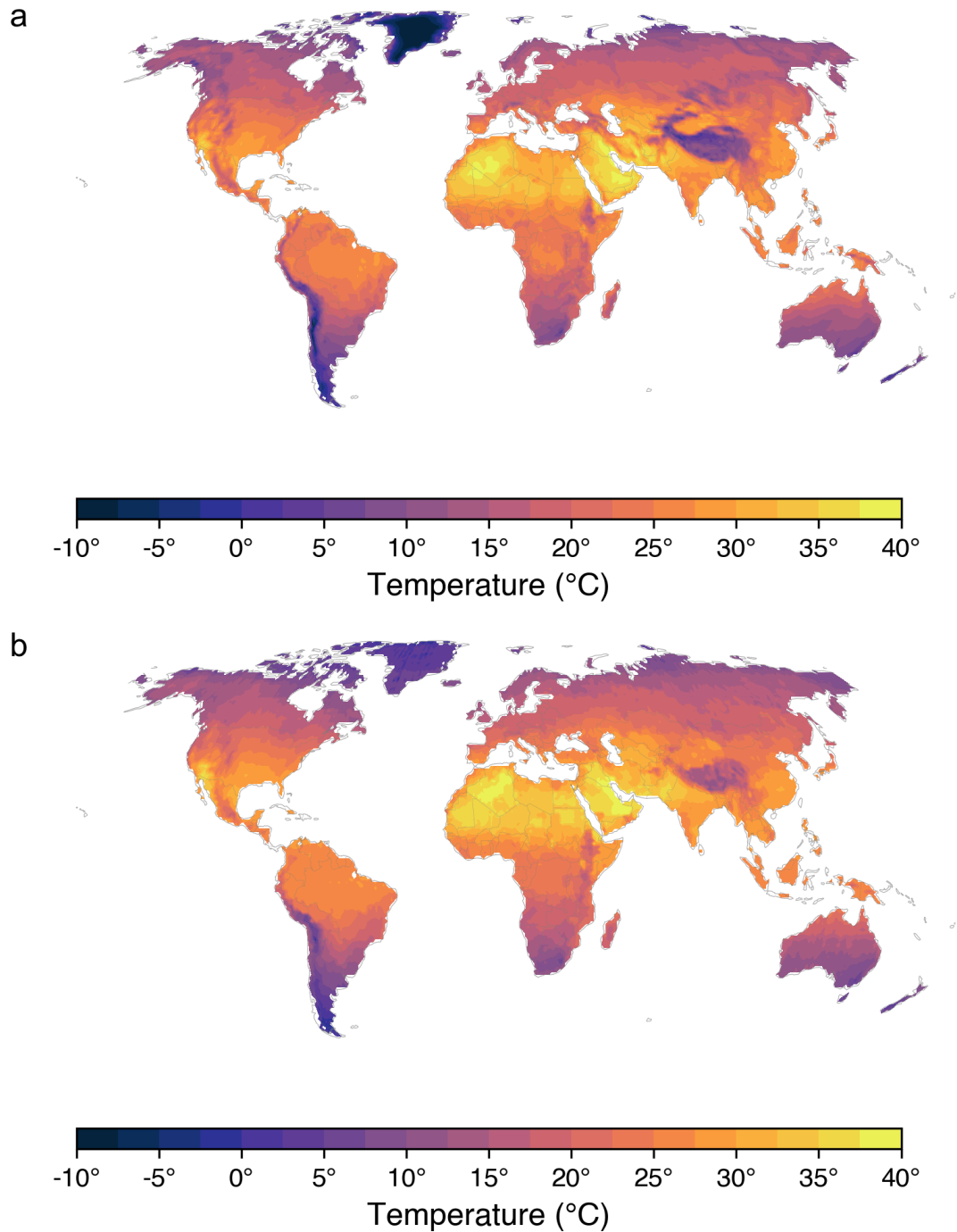

Figure 1. Mean July 2-m air temperature (1991-2020). (a) ERA5 reanalysis. (b) LLM GPT-5 predictions. Both panels are shown on a 1° grid.

Larger models within the same family generally exhibited lower errors. For instance, within the GPT-5 family, GPT-5 (RMSE 3.49 °C) outperformed GPT-5-mini (3.68 °C) and GPT-5-nano (5.24 °C). Similarly, Gemma 27B (4.49 °C) surpassed Gemma 12B (5.16 °C). Notably, models from the same family preserved the sign of their bias, with all GPT-5 variants showing cold biases and all Gemma variants showing warm biases. Among the best-performing configurations, GPT-5 (temp=0.3), GPT-5-mini (temp=0.3), GPT-OSS 120B (temp=0.3), and Mistral-small-3.1 24B (temp=0.3) all achieved RMSEs below 4.1 °C. The specialized climate model eci-io-climategpt 70B showed the poorest performance (RMSE 10.20 °C), suggesting that domain-specific fine-tuning on textual climate content does not necessarily improve numeric climate recall.

Bias maps reveal spatially coherent error patterns (Figure 2). GPT-5 exhibited cold biases in central Asia and parts of South America, with notable warm biases in Greenland and parts of the Sahara (Figure 2a). GPT-OSS 120B showed similar regional structures but with opposite signs in some areas (Figure 2b). Mistral-small-3.1 24B and Gemma 27B displayed warm biases across much of the Northern Hemisphere mid-latitudes and cold biases in parts of the Southern Hemisphere (Figures 2c,d). High-elevation regions such as the Himalayas, Tibetan Plateau, and Greenland exhibited particularly large errors across all models, indicating difficulty encoding temperature-elevation relationships. These mountainous areas are also known to show large biases in contemporary physical climate models [19], suggesting a common challenge in representing complex topography at relatively coarse resolution. In fact, our analysis shows that most biases appear in areas with altitudes exceeding 1500 m for nearly all models. To quantify this altitude dependence, we divided results into nine elevation bins (Table 2). At low elevations (0-500 m, 60.5% of land points), RMSE ranged from 1.93-3.13 °C across the four best-performing models, with biases near zero for GPT-5 (-0.14 °C) and Mistral-small-3.1 24B (-0.02 °C), but reaching -1.39 °C for GPT-OSS 120B and +0.23 °C for Gemma 27B. Performance degraded systematically with altitude: RMSE increased to 3.11-4.16 °C at 1000-1500 m, then rose sharply to 5.23-6.20 °C at 1500-2000 m. Above 2500 m (4.3% of points), errors became severe (RMSE >7 °C, biases +5 to +13 °C). Notably, 90% of locations lie below 1500 m where performance is markedly better, indicating that altitude-stratified metrics are more informative than global averages.

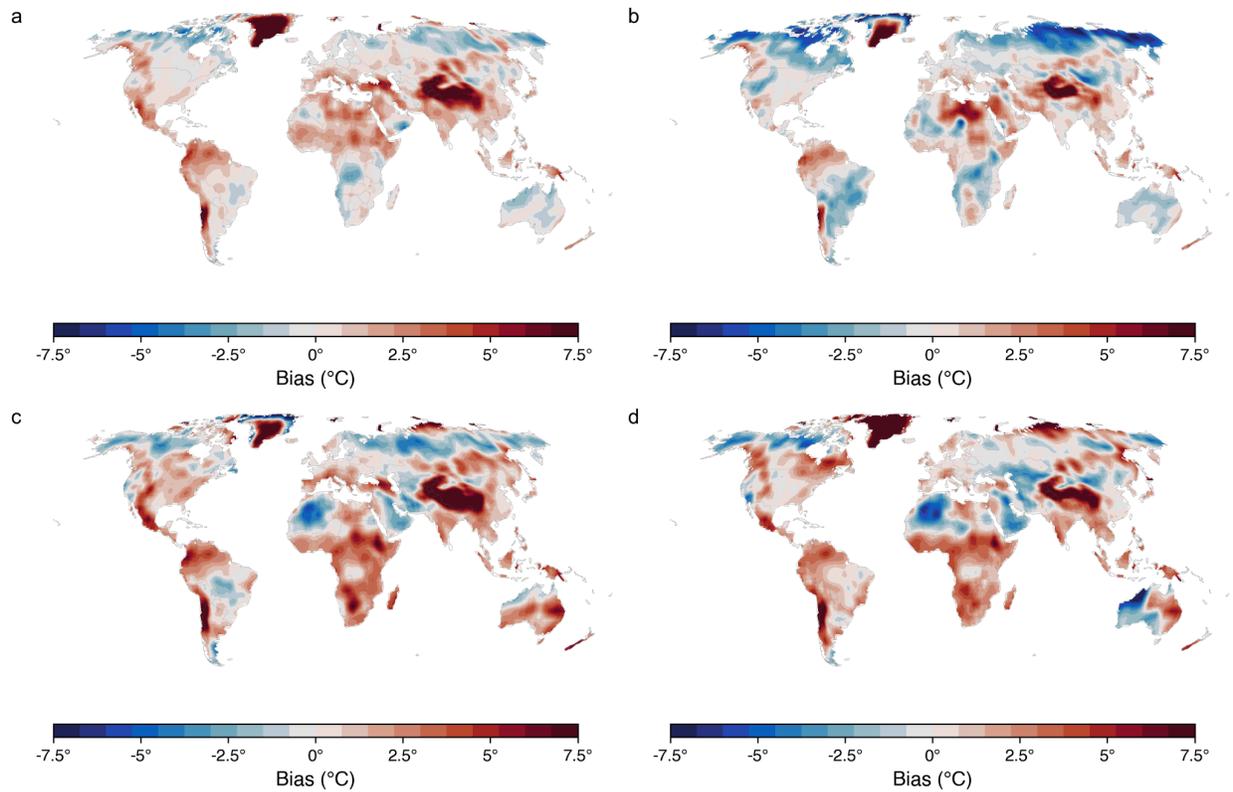

Figure 2. Bias in mean July 2-m air temperature (1991-2020), defined as LLM predictions minus ERA5 reanalysis. (a) GPT-5, (b) GPT-OSS 120B, (c) Mistral-small-3.1 24B, and (d) Gemma 27B. All simulations were run with sampling temperature set to 0.3. Bias was calculated using a neighborhood (radius = 2) scheme (see Methods).

Scatter plots of predicted versus observed temperatures (Figure 3) show strong overall correlation but systematic deviations. GPT-OSS 120B (Figure 3b) achieved near-unity slope (1.012) with relatively tight clustering around the 1:1 line, whereas GPT-5 (Figure 3a), Mistral-small-3.1 24B (Figure 3c) and Gemma 27B (Figure 3d) exhibited greater scatter. At the coldest temperatures (below 10 °C), all models tended to under-predict the magnitude of cold, consistent with their difficulty in high-latitude and high-elevation regions.

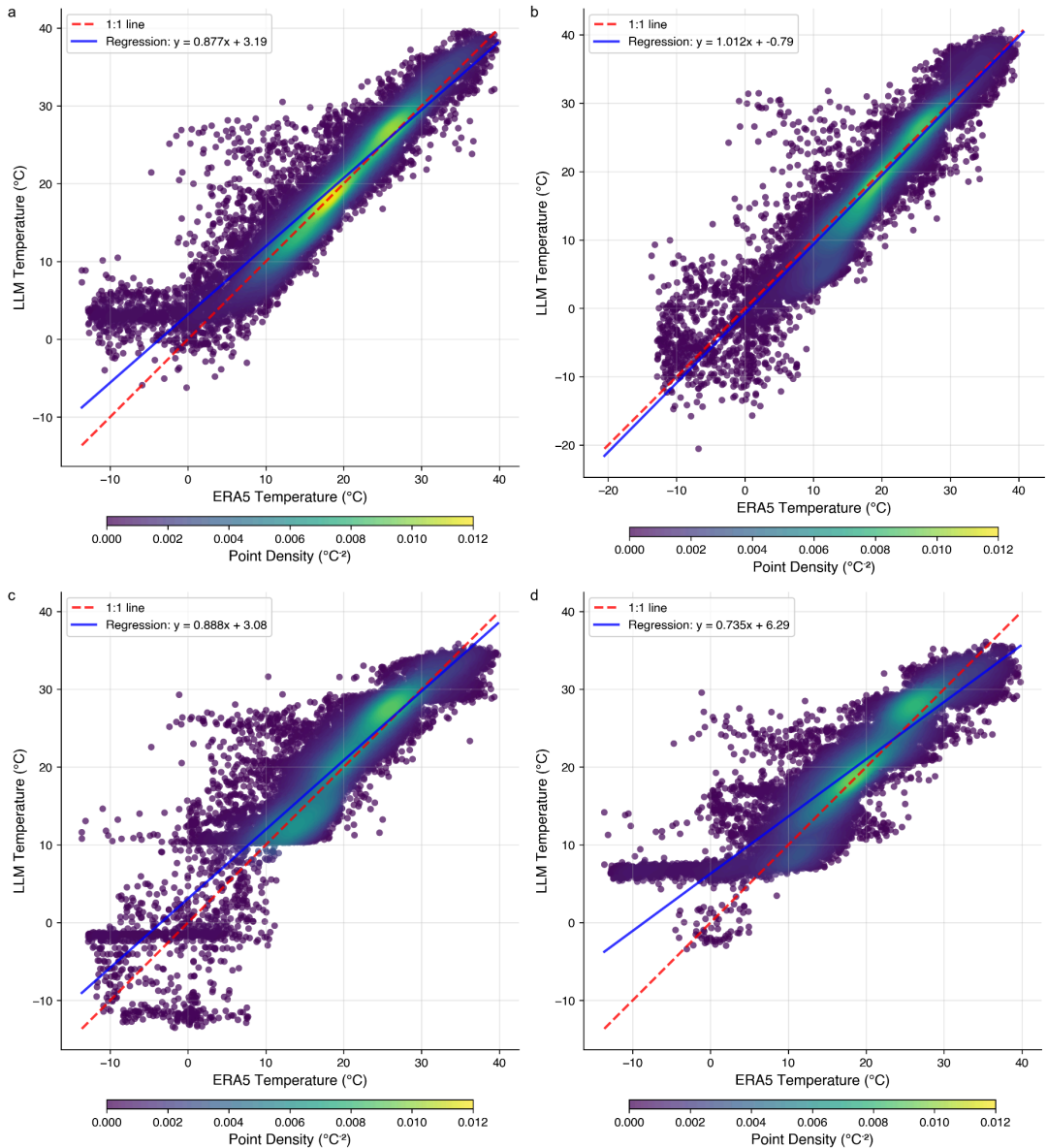

Figure 3. Scatter plots of LLM-predicted versus ERA5 2-m air temperature (1991-2020). (a) GPT-5, (b) GPT-OSS 120B, (c) Mistral-small-3.1 24B, and (d) Gemma 27B.

Sampling temperature refers to the LLM parameter controlling output randomness. We tested models at default settings and at a reduced sampling temperature of 0.3, which favors higher-probability tokens. Across most models, using a temperature of 0.3 yielded marginally improved performance. For example, GPT-OSS 120B improved from RMSE 4.03 °C (default) to 3.67 °C (temp=0.3), and GPT-5-nano showed a slight degradation from 5.24 °C to 5.29 °C.

These modest changes suggest that deterministic sampling can help models select more accurate stored values, though the effect is small and not consistent across all models.

Providing additional geographic descriptors (country, state/region, city) alongside coordinates significantly improved model accuracy. For models where we tested both conditions, RMSE increased substantially when address information was omitted (Table 1, "no address" column). Across all 13 model configurations tested with and without address information, RMSE increased by an average of 27% when geographic descriptors were omitted. For instance, GPT-5-nano RMSE rose from 5.24 °C (with address) to 6.92 °C (without address), representing a 32% increase in error, GPT-5-mini from 3.68 °C to 5.57 °C (+51%), and Gemma 27B from 4.49 °C to 7.05 °C (+57%). This result indicates that LLMs leverage recognizable place names to anchor their temperature predictions, likely because training corpora associate locations with climatic descriptions more strongly than raw coordinates. Critically, mean bias remained consistent in sign and magnitude regardless of address provision, reinforcing that omitting context increases random error rather than introducing systematic bias shifts.

We extended the evaluation to assess whether LLMs can estimate observed climate change by querying the temperature difference between two historical periods: 1950-1974 and 2000-2024. Results were poor across all tested models. Smaller models (e.g., Mistral-smal3.1 24B) produced nearly uniform warming of approximately 0.8 °C everywhere with minimal spatial variability,while GPT-5 predicted a global mean warming of 0.96 °C (Figure 4). These global mean values are reasonably close to the ERA5-derived warming of 1.05 °C for land points over this period, indicating that LLMs capture the overall magnitude of observed warming. However, models failed to reproduce the spatial distribution of temperature change: larger models such as GPT-5 exhibited some spatial variability in predicted changes but showed near-zero correlation with ERA5-derived spatial patterns of temperature trends over these periods (Figure 4).This failure to capture regional patterns can be explained by the fact that point-specific temperature trends are rarely discussed in accessible literature outside scientific publications, and hence LLMs have limited exposure to such granular temporal data during training. The results suggest that LLMs cannot generalize information to construct implicit time series for individual locations from the broader corpus of climate information. They do not automatically extract temporal patterns from scattered mentions of a location across different time periods and synthesize them into coherent trends. This contrasts sharply with their success at recalling climatological baselines, where both global statistics and spatial patterns (e.g., latitudinal gradients, topographic effects) are reasonably well reproduced. While LLMs capture the global climate change signal, they fail to represent its local and regional distribution, underscores the need for external data when temporal or spatial details of climate change dynamics are required.

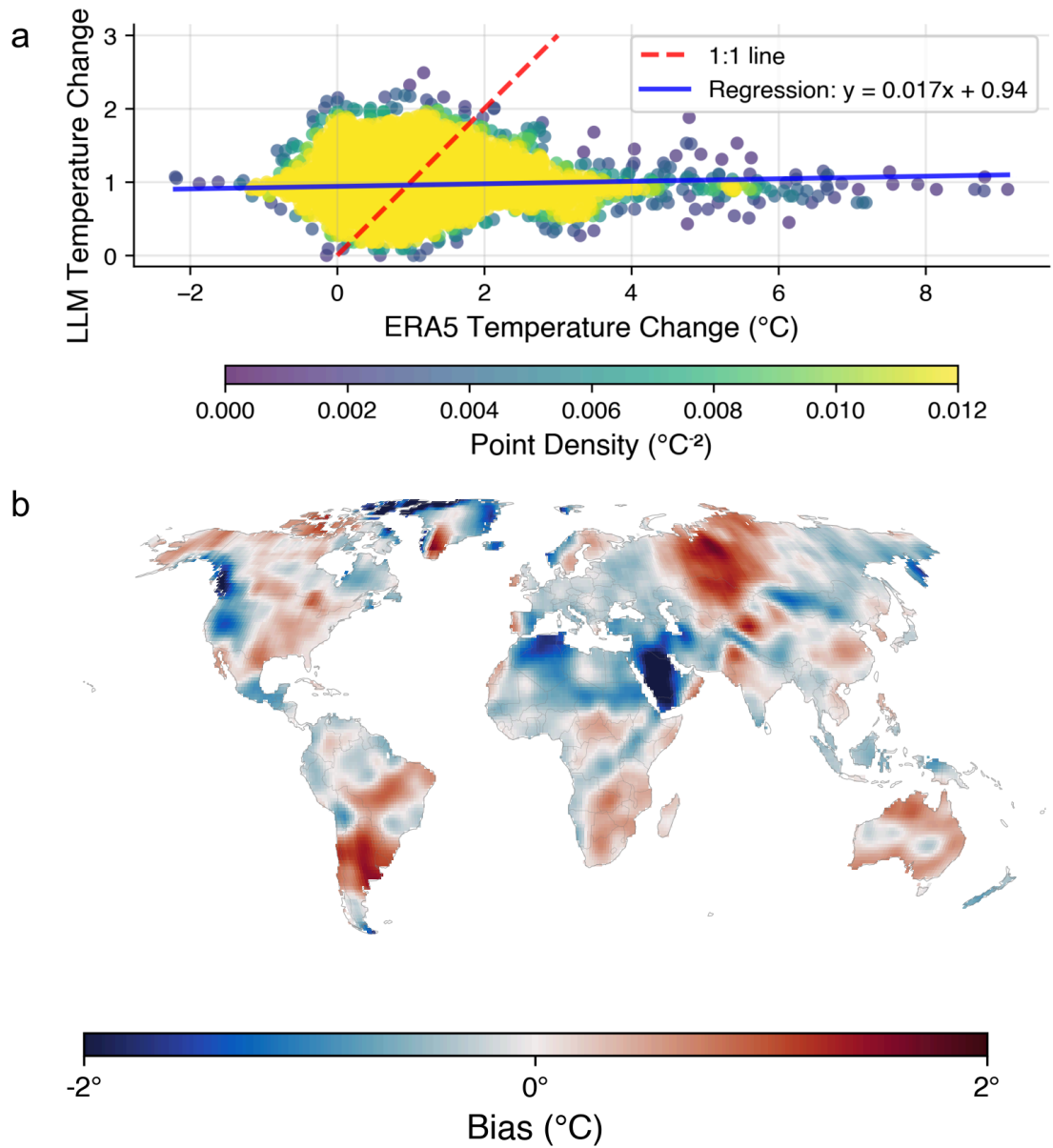

Figure 4. GPT-5 predictions of temperature change between 1950-1974 and 2000-2024. (a) Scatter plot of LLM-predicted versus ERA5-derived change in mean July 2-m air temperature. (b) Spatial bias map (LLM minus ERA5). Simulation was run with sampling temperature 0.3. Bias calculated using a neighborhood (radius = 2) scheme (see Methods).

## Discussion

Our closed-book evaluation reveals that contemporary LLMs possess non-trivial internal representations of Earth's mean climate. In qualitative terms, model-generated temperature maps capture broad expected patterns: warmer tropics and hot subtropics, cooler high latitudes and mountains. Quantitatively, however, there are significant deviations from true values, with substantial errors (RMSEs of 3 - 6 °C) and systematic regional biases that limit the reliability of unaided LLM climate recall for practical applications. The elevation-dependent analysis reveals that this limitation is concentrated: 90% of land points lie below 1500 m where errors remain moderate (RMSE 2 - 4 °C), while the remaining high-elevation locations drive much of the global error statistics.

The finding that geographic context significantly enhances accuracy has practical implications. When applications or systems query LLMs for climate information using coordinates, providing additional place names (country, region, city) substantially improves response accuracy. This suggests that climate-related applications integrating LLMs should include rich geographic descriptors alongside any coordinate data to maximize accuracy. The failure of LLMs to capture temporal changes, despite their partial success with climatological means, highlights a fundamental limitation: models encode static snapshots of knowledge but struggle with dynamic trends absent explicit training signals.

The proliferation of LLM benchmarks across domains [12-15] underscores the need for task-specific evaluation. While general-purpose benchmarks measure broad capabilities, domain benchmarks like ours provide targeted assessment of scientific knowledge recall. Our prototype focuses on a single variable and month, but the framework extends naturally. Future iterations could cover all 12 months, additional variables (precipitation, wind, extremes), and compare closed-book versus retrieval-augmented modes. Tracking performance on this benchmark as models evolve can quantify progress in parametric climate knowledge.

We emphasize that accurate numeric recall does not imply physical understanding. LLMs remain statistical pattern matchers without causal models of atmospheric processes. Their errors, while similar in magnitude to some physical model biases [19], arise from different mechanisms: incomplete or noisy training data rather than approximations to governing equations. This distinction matters when assessing reliability. For mission-critical climate applications, tool-augmented systems that ground LLM reasoning in authoritative data sources [4-6] are essential.

| Model / Configuration | RMSE / Bias (°C) | RMSE / Bias, no address (°C) |
|---|---|---|
| GPT-5, temp=0.3 | 3.49 / 0.88 | - |
| GPT-5-mini, temp=0.3 | 3.68 / -0.42 | 5.57 / -0.79 |
| GPT-5-mini | 3.68 / -0.41 | - |
| GPT-OSS 120B, temp=0.3 | 3.67 / -0.57 | 5.29 / -0.64 |
| GPT-OSS 120B | 4.03 / -0.38 | - |

| Model | RMSE / Bias | No address RMSE / Bias |
|---|---|---|
| Mistral-small-3.1 24B, temp=0.3 | 4.07 / 1.00 | - |
| Mistral-small-3.1 24B | 4.11 / 1.00 | - |
| Gemma 27B, temp=0.3 | 4.49 / 1.34 | 7.05 / 2.17 |
| Gemma 27B | 4.58 / 1.50 | - |
| GPT-OSS 20B, temp=0.3 | 5.10 / -1.32 | 6.48 / -1.33 |
| Gemma 12B, temp=0.3 | 5.16 / 1.07 | - |
| Gemma 12B | 5.18 / 1.23 | 6.31 / 0.02 |
| GPT-5-nano, temp=0.3 | 5.29 / -0.84 | - |
| GPT-5-nano | 5.24 / -0.82 | 6.92 / -0.72 |
| GPT-OSS 20B | 5.51 / -1.37 | - |
| Gemini-2.5-flash-lite, temp=0.3 | 5.51 / -1.18 | 5.57 / -1.20 |
| Gemini-2.5-flash-lite | 5.76 / -1.16 | 5.82 / -1.19 |
| Llama3.1 8B, temp=0.3 | 5.91 / 1.75 | 5.92 / 1.76 |
| Llama3.1 8B | 5.95 / 1.64 | 5.94 / 1.64 |
| Mistral-nemo 12B, temp=0.3 | 6.93 / 0.39 | 6.93 / 0.39 |
| Mistral-nemo 12B | 7.47 / 0.47 | 7.51 / 0.47 |
| Llama2 70B | 7.99 / 3.75 | 7.94 / 4.84 |
| eci-io-climategpt 70B | 10.20 / 3.93 | - |

Table 1. Root-mean-square error (RMSE) and mean bias for LLM temperature predictions. Models are sorted by minimum RMSE. All experiments used a 1.0° global land grid with July mean temperature (1991-2020). "No address" indicates experiments where country/region/city information was withheld. temp = 0.3 denotes the sampling temperature, a parameter controlling the randomness of LLM outputs, lower values produce more deterministic predictions.

| Model Altitude | GPT-5 RMSE | Bias | oss120b RMSE | Bias | Mistral-small-3.1 24B RMSE | Bias | Gemma 27B RMSE | Bias |
|---|---|---|---|---|---|---|---|---|
| 0-500m, 60.5% | 1.93 | -0.14 | 3.07 | -1.39 | 2.97 | -0.02 | 3.13 | 0.23 |
| 500-1000m, 19.2% | 2.3 | 0.61 | 3.27 | -0.34 | 3.41 | 0.84 | 3.33 | 1.11 |
| 1000-1500m, 10.3% | 3.11 | 1.5 | 3.44 | 0.21 | 3.92 | 1.88 | 4.16 | 2.01 |
| 1500-2000m, 3.9% | 5.23 | 3.76 | 4.28 | 0.89 | 4.81 | 3.08 | 6.2 | 4.07 |
| 2000-2500m, 2.3% | 8.49 | 7.27 | 5.58 | 3.02 | 6.95 | 5.76 | 10.29 | 8.84 |
| 2500-3000m, 1.4% | 11.66 | 10.64 | 7.37 | 5.12 | 8.64 | 7.69 | 14.14 | 12.81 |
| 3000-3500m, 0.6% | 12.44 | 10.92 | 9.45 | 6.95 | 11.74 | 10.42 | 14.04 | 12.34 |
| 3500-4000m, 0.5% | 11.24 | 8.58 | 10.29 | 5.41 | 12.69 | 10.97 | 11.53 | 9.68 |
| 4000m+, 1.5% | 11.29 | 9.47 | 9.47 | 5.68 | 13.88 | 12.78 | 11.59 | 9.18 |

Table 2. Model performance across elevation bins. Root-mean-square error (RMSE) and mean bias for July mean temperature predictions (1991-2020) at different altitudes. Percentages indicate the fraction of land grid points within each altitude range. All models were run with sampling temperature 0.3.

## Methods

We generated a global grid at 1° latitude-longitude resolution spanning longitudes from -180° to 180° and latitudes from -60° to 85°. From this grid, we selected 15,395 land points, excluding ocean cells. For each grid point, we extracted from ERA5 reanalysis [22] monthly-mean July 2-m air temperature for the period 1991-2020 to serve as ground truth. We used a geographic reverse-geocoding service to associate each coordinate with the nearest country, state/region, and city when available [23].

We queried models using with standardized prompts (Box 1) that requested only a numeric answer in degrees Celsius. No few-shot examples or chain-of-thought prompting was used. To assess variability, we queried each model exactly 10 times per location. If any individual request failed to return a valid numeric response, we re-queried that specific request up to 3 additional times. Non-numeric or out-of-range responses (<-90 °C or >60 °C) were flagged during this process. The final prediction for each location was taken as the mean of the 10 successful responses. We tested models at two sampling temperature settings: temperature = 0 (deterministic) and temperature = 0.3 (slightly stochastic).

For evaluation metrics reported in Table 1, we computed root-mean-square error (RMSE) and mean bias directly between model predictions and ERA5 values across all land points without area weighting. This provides a straightforward assessment of point-wise accuracy. For spatial visualization in bias maps (Figure 2), we employed neighborhood-averaged bias calculations following standard practice in climate model evaluation. Specifically, we assessed spatial consistency using a neighborhood scheme: for each land grid point, metrics were computed over a 5×5 window (radius = 2 grid cells) with latitude clipping, longitudinal wraparound, and a requirement of at least 6 valid neighbors [20]. Windows not meeting the neighbor threshold were set to missing, and ocean cells were excluded. This spatial smoothing highlights coherent regional error patterns and accounts for the varying area represented by grid cells at different latitudes.

For scatter plots (Figure 3), we computed point density using a Gaussian kernel density estimator with default Scott bandwidth [21]. For histograms (Figure 4), we binned the temperature differences (LLM minus ERA5) and overlaid a fitted normal distribution.

Claude Code (Anthropic) assisted in generating portions of the analysis code under full human oversight and validation.

| You are a climate data expert. Given the location coordinates and address information below, provide the mean July temperature for the period 1991-2020.<br><br>Location Information:<br>- Longitude: {{longitude}}<br>- Latitude: {{latitude}}<br>- Country: {{country}}<br>- State/Region: {{state}}<br>- City: {{city}}<br><br>Provide ONLY the mean July temperature at 2m above surface (°C) for this location for the climatological period 1991-2020.<br><br>IMPORTANT: Return ONLY a single number (float) representing the mean July temperature in Celsius. No text, no JSON, just the number.<br><br>Example: 25.4 | You are a climate data expert. Given the location coordinates and address information below, calculate the change in mean July temperature between the periods 1950-1974 and 2000-2024.<br><br>Location Information:<br>- Longitude: {{longitude}}<br>- Latitude: {{latitude}}<br>- Country: {{country}}<br>- State/Region: {{state}}<br>- City: {{city}}<br><br>Provide ONLY the temperature difference (2000-2024 minus 1950-1974) at 2 m above surface (°C) for this location.<br><br>IMPORTANT: Return ONLY a single number (float) representing the temperature difference in Celsius. No text, no JSON, just the number.<br><br>Example: 1.8 |
|---|---|

Box 1. Prompt templates. Left: prompt for climatological baseline (1991-2020). Right: prompt for temporal change (temperature difference, 2000-2024 minus 1950-1974).

## Data and code availability

Code is available at https://github.com/CliDyn/geo_benchmark

ERA5 monthly averaged data on single levels are available from the Copernicus Climate Data Store [22] (https://cds.climate.copernicus.eu/datasets/reanalysis-era5-single-levels-monthly-means) (accessed 07 July 2025). Contains modified Copernicus Climate Change Service information

2025. Neither the European Commission nor ECMWF is responsible for any use that may be made of the Copernicus information or data it contains.

Acknowledgements

This work was supported by the European Union's Destination Earth Initiative and relates to tasks entrusted by the European Union to the European Centre for Medium-Range Weather Forecasts implementing part of this Initiative with funding by the European Union. This work is also supported by projects S1: Diagnosis and Metrics in Climate Models of the Collaborative Research Centre TRR 181 "Energy Transfer in Atmosphere and Ocean", funded by the Deutsche Forschungsgemeinschaft (DFG, German Research Foundation, project no. 274762653). It is also supported by the EERIE project (Grant Agreement No 101081383), funded by the European Union.

# References

[1] Kandpal, N., Deng, H., Roberts, A., Wallace, E. & Raffel, C. Large language models struggle to learn long-tail knowledge. In *Proc. 40th Int. Conf. Machine Learning (ICML 2023)*, PMLR 202, 15696-15707 (2023).

[2] Yuan, J., Pan, L., Hang, C.-W., Guo, J., Jiang, J., Min, B., Ng, P. & Wang, Z. Towards a holistic evaluation of LLMs on factual knowledge recall. *arXiv* 2404.16164 (2024). https://doi.org/10.48550/arXiv.2404.16164

[3] Wang, Y., Chen, Y., Wen, W., Sheng, Y., Li, L. & Zeng, D. D. Unveiling factual recall behaviors of large language models through knowledge neurons. In Proc. 2024 Conf. Empirical Methods in Natural Language Processing (eds Al-Onaizan, Y., Bansal, M. & Chen, Y.-N.) 7388-7402 (Association for Computational Linguistics, Miami, Florida, USA, 2024). https://doi.org/10.18653/v1/2024.emnlp-main.420

[4] Vaghefi, S. A., Mansoor, W., McKee, S., Gao, R., Grabow, C., Chakraborty, A., Luo, J., Moss, R. H., Shapiro, C., Braneon, C., van den Bergh, J., Calvin, K. V., Fu, J. S., Jiang, Y., Kang, J. E., Kumar, R., Lebling, K., Mistry, M., Morgan, M. G., Rahim, A., Singh, R., Tan, J., Tavoni, M., Winstral, A., Yuan, R., Zhang, H., Zheng, H., Diffenbaugh, N. S. & Rolnick, D. ChatClimate: Grounding conversational AI in climate science. Commun. Earth Environ. 4, 480 (2023). https://doi.org/10.1038/s43247-023-01084-x

[5] Koldunov, N. & Jung, T. Local climate services for all, courtesy of large language models. *Commun. Earth Environ.* 5, 13 (2024). https://doi.org/10.1038/s43247-023-01199-1

[6] Kuznetsov, I., Jost, A. A., Pantiukhin, D., Shapkin, B., Jung, T. & Koldunov, N. Transforming climate services with LLMs and multi-source data integration. *npj Climate Action* (in press, 2025).

[7] Bulian J. et al. Assessing large language models on climate information. In Proc. 41st Int. Conf. Machine Learning (eds Salakhutdinov R. et al.) Proc. Mach. Learn. Res. 235, 4884–4935 (2024).

[8] Min, S., Krishna, K., Lyu, X., Lewis, M., Yih, W.-t., Koh, P., Iyyer, M., Zettlemoyer, L. & Hajishirzi, H. FAactScore: Fine-grained atomic evaluation of factual precision in long form text generation. In Proc. 2023 Conf. Empirical Methods in Natural Language Processing (eds Bouamor, H., Pino, J. & Bali, K.) 12076-12100 (Association for Computational Linguistics, Singapore, 2023). https://doi.org/10.18653/v1/2023.emnlp-main.741

[9] Zhang, Y., Ren, X., Xu, C., Zhang, Y., Wu, Z., Wang, L., Tang, X., Song, K., Zhu, C. & Zhou, M. Siren's song in the AI ocean: A survey on hallucination in large language models. arXiv 2309.01219 (2023). https://doi.org/10.48550/arXiv.2309.01219

[10] Tonmoy, S. M., Kumar, S., Hassan, S., Mahmud, M., Ahmed, S., Sen, A., Zaman, K., Roy, A., Ahmed, R., Barman, S., Roy, T., Rakshit, S., Islam, S. & Mahmud, H. A comprehensive survey of hallucination mitigation techniques in large language models. arXiv 2401.01313 (2024). https://doi.org/10.48550/arXiv.2401.01313

[11] Chang, H., Park, J., Ye, S., Yang, S., Seo, Y., Chang, D., & Seo, M. How do large language models acquire factual knowledge during pretraining? In Proc. 38th Conf. Neural Information Processing Systems (NeurIPS 2024) (NeurIPS, 2024). Preprint on arXiv:2406.11813. https://doi.org/10.48550/arXiv.2406.11813

[12] Liang, P., Bommasani, R., Lee, T., Tsipras, D., Li, X., Zhang, X. et al. Holistic evaluation of language models. arXiv preprint arXiv:2211.09110 (2022). https://doi.org/10.48550/arXiv.2211.09110

[13] Chiang, W.-L., Zheng, L., Zhuang, S., Wallace, E., Shen, Z., Savarese, S., Finn, C. & Zhang, T. Chatbot Arena: An open platform for evaluating LLMs by human preference. In Proc. Int. Conf. on Machine Learning (ICML 2024) (ICML, 2024). Preprint on arXiv:2403.04132. doi:10.48550/arXiv.2403.04132

[14] Zheng, L., Chiang, W.-L., Zhuang, S., Shen, Z., Savarese, S., Finn, C. & Zhang, T. Judging LLM-as-a-judge with MT-Bench and Chatbot Arena. In Proc. 2023 Conf. Neural Information Processing Systems (NeurIPS 2023) (NeurIPS, 2023). Preprint on arXiv:2306.05685 (2023). doi:10.48550/arXiv.2306.05685

[15] Srivastava, A. et al. Beyond the imitation game: Quantifying and extrapolating the capabilities of language models. *arXiv* 2206.04615 (2022). https://doi.org/10.48550/arXiv.2206.04615

[16] Feng, K. et al. SciKnowEval: Evaluating multi-level scientific knowledge of large language models. *arXiv* 2406.09098 (2024). https://doi.org/10.48550/arXiv.2406.09098

[17] Sun, L. et al. SciEval: A multi-level large language model evaluation benchmark for scientific research. *arXiv* 2308.13149 (2023). https://doi.org/10.48550/arXiv.2308.13149

[18] Thulke, D. et al. ClimateGPT: Towards AI synthesizing interdisciplinary research on climate change. *arXiv* 2401.09646 (2024). https://doi.org/10.48550/arXiv.2401.09646

[19] Eyring, V. et al. Overview of the Coupled Model Intercomparison Project Phase 6 (CMIP6) experimental design and organization. *Geosci. Model Dev.* 9, 1937-1958 (2016). https://doi.org/10.5194/gmd-9-1937-2016

[20] Gilleland, E. et al. Intercomparison of spatial forecast verification methods. *Wea. Forecasting* 24, 1416-1430 (2009). https://doi.org/10.1175/2009WAF2222269.1

[21] Scott, D. W. *Multivariate Density Estimation: Theory, Practice, and Visualization* 2nd edn (Wiley, 2015). https://doi.org/10.1002/9781118575574

[22] Copernicus Climate Change Service. ERA5 monthly averaged data on single levels from 1940 to present. Copernicus Climate Change Service (C3S) Climate Data Store (CDS) https://doi.org/10.24381/cds.f17050d7 (2023).

[23] Nominatim. Nominatim (OpenStreetMap Foundation) https://nominatim.org/ (accessed 11 March 2025).